\pdfoutput=1

\documentclass[11pt]{article}

\usepackage{acl}

\usepackage{times}
\usepackage{latexsym}
\usepackage{CJKutf8}
\usepackage{amsmath}
\usepackage{verbatim}
\usepackage{booktabs}
\usepackage{bm}
\usepackage{amssymb}
\usepackage{pifont}
\usepackage{stfloats}
\usepackage{multicol}
\usepackage{float}
\usepackage{graphicx}
\usepackage{subfigure}
\usepackage{multirow}
\usepackage{array}
\newcommand{\PreserveBackslash}[1]{\let\temp=\\#1\let\\=\temp}
\newcolumntype{C}[1]{>{\PreserveBackslash\centering}p{#1}}
\newcolumntype{R}[1]{>{\PreserveBackslash\raggedleft}p{#1}}
\newcolumntype{L}[1]{>{\PreserveBackslash\raggedright}p{#1}}
\usepackage{diagbox}
\usepackage{arydshln}
\usepackage{booktabs}
\usepackage[ruled,linesnumbered,lined,commentsnumbered]{algorithm2e}
\usepackage{mathtools}
\usepackage{amsfonts}
\usepackage{bbm}
\usepackage{url}

\makeatletter
\def\hlinew#1{%
  \noalign{\ifnum0=`}\fi\hrule \@height #1 \futurelet
   \reserved@a\@xhline}
\makeatother

\usepackage[T1]{fontenc}

\usepackage[utf8]{inputenc}

\usepackage{microtype}

\title{Gaussian Multi-head Attention for Simultaneous Machine Translation}

\author{Shaolei Zhang \textsuperscript{\rm 1,2},
    Yang Feng \textsuperscript{\rm 1,2}\thanks{ $\;\;$Corresponding author: Yang Feng. $\;\;\;\;\;\;\;\;\;\;\;\;\;\;\;\;\;\;\;\;\;$ Code is available at: \url{https://github.com/ictnlp/GMA}} \\
        \textsuperscript{\rm 1}{Key Laboratory of Intelligent Information Processing} \\ Institute of Computing Technology, Chinese Academy of Sciences (ICT/CAS) \\
    { \textsuperscript{\rm 2} {University of Chinese Academy of Sciences, Beijing, China}} \\
     \texttt{\{zhangshaolei20z, fengyang\}@ict.ac.cn}  }

\begin{document}
\maketitle
\begin{abstract}

Simultaneous machine translation (SiMT) outputs translation while receiving the streaming source inputs, and hence needs a policy to determine where to start translating. The alignment between target and source words often implies the most informative source word for each target word, and hence provides the unified control over translation quality and latency, but unfortunately the existing SiMT methods do not explicitly model the alignment to perform the control. In this paper, we propose \emph{Gaussian Multi-head Attention} (\emph{GMA}) to develop a new SiMT policy by modeling alignment and translation in a unified manner. For SiMT policy, GMA models the aligned source position of each target word, and accordingly waits until its aligned position to start translating. To integrate the learning of alignment into the translation model, a Gaussian distribution centered on predicted aligned position is introduced as an alignment-related prior, which cooperates with translation-related soft attention to determine the final attention. Experiments on En$\rightarrow$Vi and De$\rightarrow$En tasks show that our method outperforms strong baselines on the trade-off between translation and latency.

\end{abstract}

\section{Introduction}

Simultaneous machine translation (SiMT) \cite{gu-etal-2017-learning,ma-etal-2019-stacl,Arivazhagan2019}, which outputs translation before receiving the complete source sentence, is mainly used for streaming translation tasks, such as simultaneous interpretation, live broadcast and online translation. Different from full-sentence machine translation which waits for the complete source sentence, SiMT requires a policy to determine where to start translating when given the streaming inputs. The SiMT policy has to trade off between translation quality and latency and an ideal one should wait for the right number of source words, which are sufficient but not excess, until deciding to output target words \cite{Arivazhagan2019}.

For full-sentence translation, each target word is generated based on the attended source information, where each source word provides different amount of information for the target word. Among them, the most informative source word can be considered as an aligned word to the target word \cite{garg-etal-2019-jointly}. Then for SiMT, the alignment can be a good guider for the policy to determine where to start translating. For high translation quality, the SiMT policy is supposed to start translating after receiving the aligned source word to ensure enough source information for the translation. To consider low latency, the SiMT policy is expected not to wait for too many words after receiving the aligned source word. Therefore, if the alignment can be modeled in the SiMT model explicitly, translation quality and latency can be controlled in a unified manner for an ideal SiMT policy.

\begin{figure}[t]
\centering
\subfigure[Predict READ/WRITE action]{
\includegraphics[width=3in]{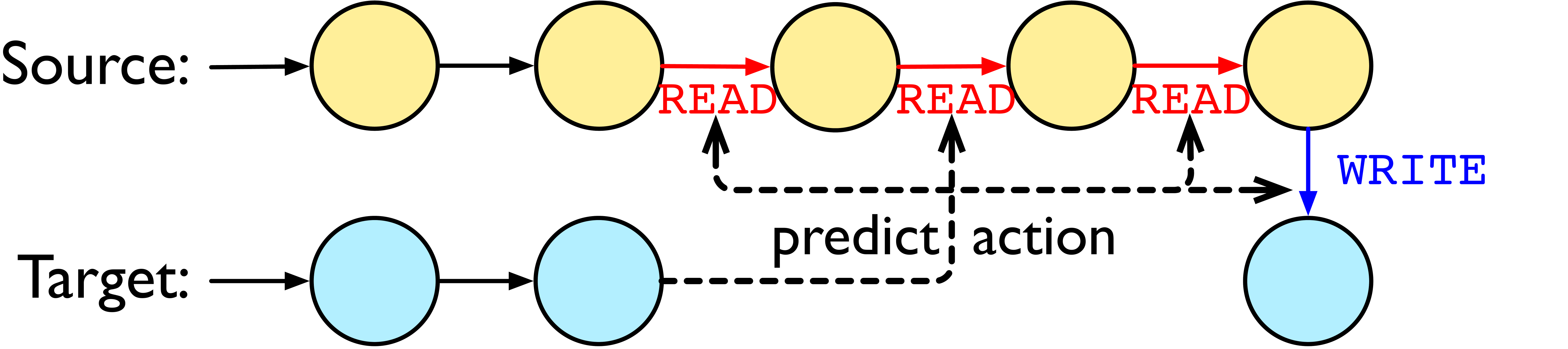}
\label{ill1}
}
\subfigure[Predict incremental step]{
\includegraphics[width=3in]{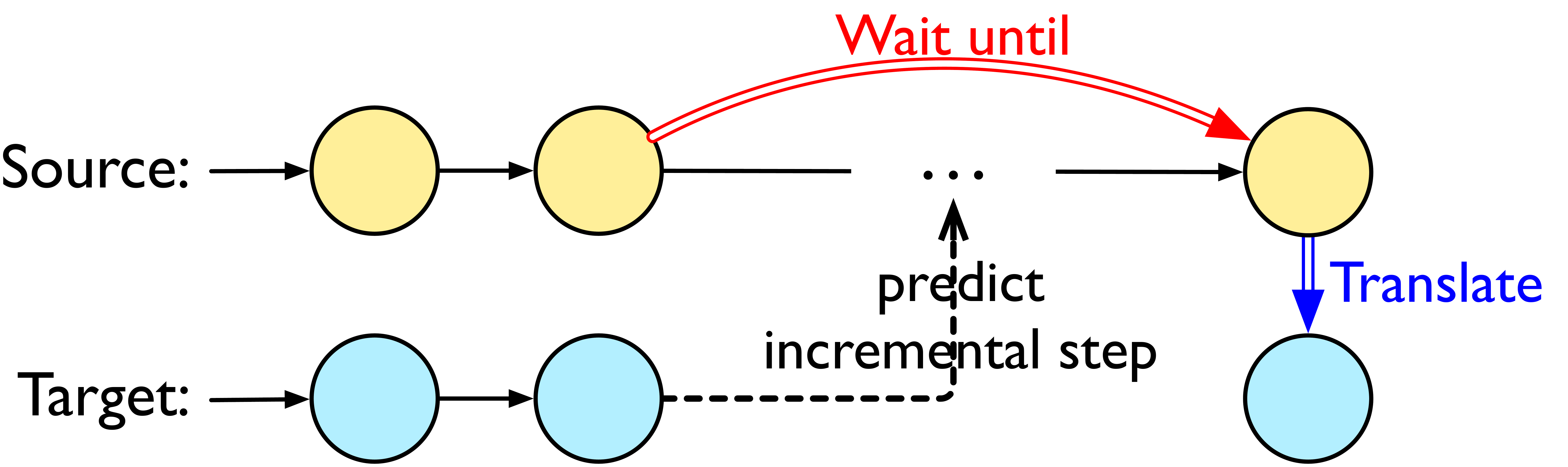}
\label{ill2}
}

\caption{Comparison diagram of previous and proposed SiMT policy. The previous policies always predict the READ/WRITE action step by step, while our method directly predicts \emph{incremental step} (i.e., number of waiting words) between two adjacent target words.}
\label{ill}
\end{figure}

However, the existing SiMT methods, mostly employing fixed or adaptive policy, do not reflect the alignment information in the modeling, which makes them trade off between translation quality and latency in a separate way. Fixed policy, such as wait-k policy \cite{ma-etal-2019-stacl}, waits for a fixed number of source words and then performs READ and WRITE operations one after another, so it is a rule-based policy and precludes alignment modeling. Adaptive policy, such as MILk \cite{Arivazhagan2019} and MMA \cite{Ma2019a}, determines READ/WRITE operations by sampling from a Bernoulli distribution, as shown in Figure \ref{ill1}, where the decisions are made independently and no relationship between the decision and translation is introduced, so it has to employ an additional loss to control the latency. Besides, some methods \cite{wilken-etal-2020-neural,arthur-etal-2021-learning} apply external ground-truth alignment as an ideal position to start translating, but the performance is inferior to MMA since separating translation and alignment.

In this paper, to explicitly involve alignments in the SiMT modeling, we propose \emph{Gaussian Multi-head Attention} (\emph{GMA}) to develop a SiMT policy with the guidance of alignments. To determine where to start translating with alignment, GMA first models the aligned source position of the current target word via predicting the relative distance from the previous aligned source position, called \emph{incremental steps}, shown in Figure \ref{ill2}. Meanwhile, a relaxation offset after the aligned position is set to allow the model to wait for some additional source inputs, thereby providing a controllable trade-off between translation quality and latency in practice. Accordingly, GMA starts translating after receiving the aligned source position and waiting for the extra relaxation offset. To jointly learn alignments (i.e., SiMT policy) and translation, a Gaussian distribution centered on predicted aligned position is introduced as a prior attention over the received source words. As a result, GMA finally uses the posterior attention for translation derived from the alignment-related Gaussian prior and translation-related soft attention. Experiments on En$\rightarrow$Vi and De$\rightarrow$En SiMT tasks show that GMA outperforms strong baselines on the trade-off between translation quality and latency.

\section{Background}

GMA is applied on the multi-head attention in Transformer \cite{NIPS2017_7181}, so we briefly introduce SiMT and the multi-head attention.

\subsection{Simultaneous Machine Translation}
In a translation task, we denote the source sentence as $\mathbf{x}\!=\!\left \{ x_{1},\cdots ,x_{J} \right \}$ and the corresponding source hidden states as $\mathbf{z}\!=\!\left \{ z_{1},\cdots ,z_{J} \right \}$ with source length $J$. The model generates a target sentence $\mathbf{y}\!=\!\left \{ y_{1},\cdots ,y_{I} \right \}$ and the corresponding target hidden states $\mathbf{s}\!=\!\left \{ s_{1},\cdots ,s_{I} \right \}$ with target length $I$. Different from the full-sentence machine translation, the source words received by SiMT model are incremental and hence the model needs to decide where to output translation.

\textbf{Output position} Define $g(i)$ \cite{ma-etal-2019-stacl} as a monotonic non-decreasing function of step $i$, to denote the number of source words received by SiMT model when translating $y_{i}$, i.e., $g(i)$ is the output position of $y_{i}$.

In SiMT, $g(i)$ is determined by the specific policy, and the probability of generating the target word $y_{i}$ is $p\left ( y_{i}\mid \mathbf{x}_{\leq g(i)},\mathbf{y}_{< i} \right )$, where $\mathbf{x}_{\leq g(i)}$ is first $g(i)$ source words and $\mathbf{y}_{< i}$ is previous target words. Therefore, the decoding probability of $\mathbf{y}$ is calculated as:
\begin{gather}
    p(\mathbf{y}\mid \mathbf{x})=\prod_{i=1}^{\left | \mathbf{y} \right |}p\left ( y_{i}\mid \mathbf{x}_{\leq g(i)},\mathbf{y}_{< i} \right )
\end{gather}

\subsection{Multi-head Attention}

Multi-head attention \cite{NIPS2017_7181} contains multiple attention heads, where each attention head performs scaled dot-product attention. Our method is based on the cross-attention, where the queries are the target hidden states $\mathbf{s}$, the keys and values both come from the source hidden states $\mathbf{z}$. The soft attention weight $\alpha _{ij}$ is calculated as:
\begin{gather}
\alpha_{ij} = \mathrm{Softmax}(\frac{Q\!\left ( s_{i} \right )K\!\left ( z_{j} \right )^{\top}}{\sqrt{d_{k}}})\label{eq2}
\end{gather}
where $Q\!\left ( \cdot  \right )$ and $K\!\left ( \cdot  \right )$ are projection functions from the input space to the query and key space respectively, and $d_{k}$ is the dimension of inputs. Then the context vector $c_{i}$ is calculated as:
\begin{gather}
c_{i}=\sum_{j=1}^{J} \alpha _{ij}V\!\left ( z_{j} \right )
\end{gather}
where $V\!\left ( \cdot  \right )$ is a projection function to value space.

\section{The Proposed Method}
\label{sec:method}
\begin{figure*}[t]
\centering
\includegraphics[width=6in]{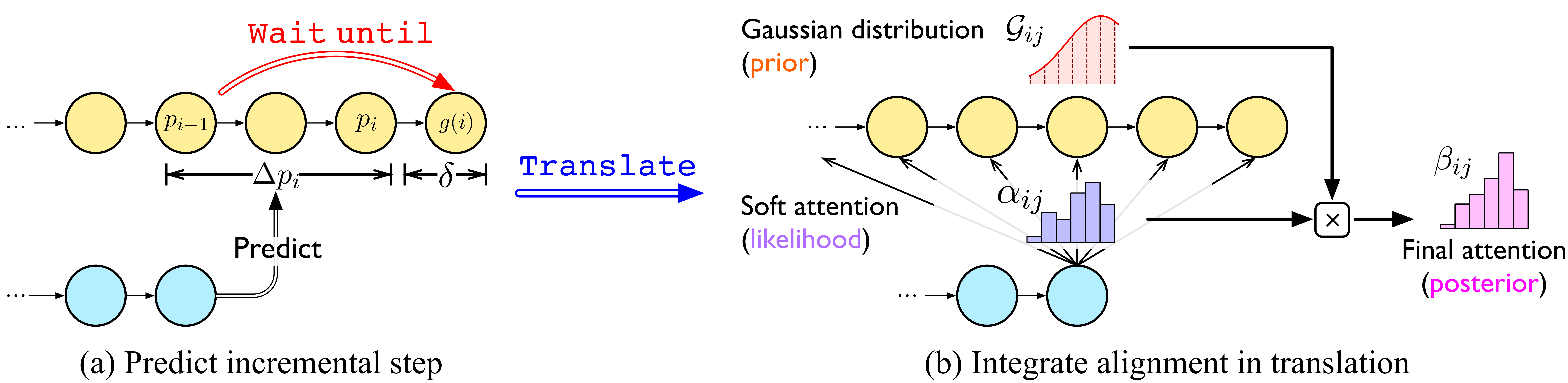}
\caption{The architecture of GMA. (a) GMA first models the aligned source position of the current target word via predicting incremental step $\Delta p_{i}$, and then waits until the aligned position to start translating. (b) To integrate the learning of alignment (which determines the latency) in translation, we introduce a Gaussian distribution centered on aligned source position as alignment-related prior probability, which is multiplied with soft attention (likelihood probability) to get final attention distribution (posterior probability).}
\label{model}
\end{figure*}

The architecture of GMA is shown in Figure \ref{model}. For SiMT policy, GMA predicts the aligned source position of the current target word, and accordingly determines the output position. To integrate the learning of SiMT policy within the translation, we introduce a Gaussian prior centered on the predicted aligned position, which is multiplied with soft attention (Eq.(\ref{eq2})) to get final attention distribution. Due to the unimodality of Gaussian prior, it enables the model to learn the position that gets the highest soft attention (i.e., alignment), thereby developing a reasonable SiMT policy. Besides, since the Gaussian prior is continuous and differentiable, it can be integrated into the translation model directly and adjusted with the learning of translation.

\subsection{Alignment-Guided SiMT Policy}

\textbf{Alignments prediction} To develop a SiMT policy with alignments, GMA first predicts the aligned source position of the current target word. Due to the incrementality of streaming inputs in SiMT, it is unstable to directly predict the absolute position of the aligned source word. Instead, we predict the relative distance from the previous aligned source position, called \emph{incremental step}.

Formally, we denote the aligned source position of the $i^{th}$ target word as $p_{i}\! \in\! \left [1,J  \right ] $ and the incremental step as $\Delta p_{i} \!\in\! \left ( 0,+\infty  \right ) $. Therefore, the aligned source position $p_{i}$ is calculated as:
\begin{gather}
    p_{i}=\left\{\begin{matrix}
1 & i=0\\ 
p_{i-1}+\Delta p_{i} & i>0
\end{matrix}\right. \label{eq7}
\end{gather}
where we set the initial aligned position $p_{0}$ to the first source word, and the incremental step $\Delta p_{i}$ is predicted through a multi-layer perceptron (MLP) based on the previous target hidden state $s_{i-1} $:
\begin{gather}
    \Delta p_{i}=\mathrm{exp}\left (\mathbf{ V_{\!p}}^{\top }\mathrm{tanh}\!\left ( \mathbf{W_{\!p}}\; Q\!\left ( s_{i-1} \right )\right )\right )\label{eq8}
\end{gather}
where $\mathbf{W_{\!p}}$, $\mathbf{ V_{\!p}}$ are learnable parameters of MLP.

\begin{algorithm}[t]
\caption{SiMT Policy of GMA}\label{algorithm}
  \SetKwData{Training}{Training}\SetKwData{This}{this}\SetKwData{Up}{up}
  \SetKwFunction{getInputTestLagging}{getInputTestLagging}\SetKwFunction{SampleFrom}{SampleFrom}
  \SetKwInOut{Input}{Input}\SetKwInOut{Output}{Output}
  \Input{Streaming inputs $\mathbf{x}$ (incremental),  Initial aligned position $p_{0}=1$, $i=1$, $y_{0}=\left \langle \mathrm{BOS} \right \rangle$}
  \Output{Target outputs $\mathbf{y}$}
  \BlankLine
  \While{$y_{i-1}\neq\left \langle  \mathrm{EOS} \right \rangle$}{
    calculate \emph{incremental step} $\Delta p_{i}$ as Eq.(\ref{eq8}) \\
    $p_{i}\leftarrow p_{i-1}+\Delta p_{i}$ \\
    $g(i)=\left \lfloor p_{i}+\delta  \right \rfloor$ \\
    \eIf(\tcp*[f]{$\!\!\!\triangleright $Wait}){$g(i)>\left | \mathbf{x} \right |$}{
      $\!\!$\textbf{Wait Until} receive $g(i)$ source words \\
      $\!\!$\textbf{continue}
    }
    (\tcp*[f]{$\!\!\!\triangleright $Translate}){
      $\!\!$\textbf{Translate} $y_{i}$ with $\mathbf{x} _{\leq g(i)}$ and $\mathbf{y}_{< i}$
    }
    $i\leftarrow i+1$

    }

\end{algorithm}

\textbf{SiMT policy}
Besides the predicted aligned position $p_{i}$, we also introduce a \emph{relaxation offset} $\delta$ to allow the model to wait for some additional source inputs, thereby providing a controllable trade-off between translation quality and latency in practice. Specifically, the output position $g(i)$ (i.e., wait for the first $g(i)$ source words and then translate the $i^{th}$ target word) is calculated as:
\begin{gather}
    g(i)=\left \lfloor p_{i}+\delta  \right \rfloor \label{eq9}
\end{gather}
where $\left \lfloor \cdot  \right \rfloor$ is a floor operation. In our experiments, relaxation offset $\delta$ is a hyperparameter we set to obtain the translation quality under different latency. Overall, the SiMT policy is shown in Algorithm \ref{algorithm}.
 
\subsection{Integrating Alignment in Translation}

To jointly learn the SiMT policy (i.e., aligned positions which determine latency) with translation, we weaken the attention of source words far away from the predicted aligned position in advance, thereby forcing the model to move the predicted aligned position to the source word that is most informative for translation (i.e., with the highest soft attention).

To this end, for $i^{th}$ target word, we introduce a Gaussian distribution $\bm{\mathcal{G}_{i}}$ centered on aligned position $p_{i}$ as the prior probability, calculated as:
\begin{gather}
\!\!\mathcal{G}_{ij}\!=\!\left\{\begin{matrix}
\frac{1}{\sqrt{2\pi} \sigma_{i} } \mathrm{exp}\left (\! -\frac{\left ( j-p_{i} \right )^{2}}{2\sigma_{i}^{2} } \right ) & \mbox{if } j\!\leq\! g(i) \\ 0
 & \mbox{otherwise}
\end{matrix}\right. 
\end{gather}
where the Gaussian distribution is limited in first $g(i)$ source words. $\sigma _{i}$ is the variance used to control the attenuation degree of the prior probability as away from the aligned position. To prevent the prior probability of the furthest source word from being too small, we set $\sigma _{i}={p_{i}}/{2}$ according to the ``two-sigma rule'' \cite{2sigma}. We will compare the performance of different settings of prior probability in Sec.\ref{sec:ab}. Note that since the source position is discrete, we normalize the Gaussian distribution with $\mathcal{G}_{ij}/\sum_{k=1}^{g(i)}\mathcal{G}_{ik}$.

Given the prior probability $\mathcal{G}_{ij}$ and soft attention $\alpha_{ij}$ (calculated as Eq.(\ref{eq2})), which is considered as likelihood probability, we calculate the posterior probability $ \widehat{\beta}  _{ij}$ and normalize it as the final attention distribution $\beta  _{ij}$:

\begin{align}
    \widehat{\beta}  _{ij}=&\alpha_{ij}\times \mathcal{G}_{ij}\\
    \beta  _{ij}=&\frac{\widehat{\beta}  _{ij}}{\sum_{k=1}^{g(i)}\widehat{\beta}  _{ik}}
\end{align}
Then, the context vector $c_{i}$ is calculated as:
\begin{gather}
    c_{i}=\sum_{j=1}^{g(i)}\, \beta _{ij}V\!\left ( z_{j} \right )
\end{gather}

\subsection{Adaptation to Multi-head Structure}

When GMA is integrated into the Transformer with $L$ decoder layers and $H$ attention heads per layer, if multiple heads (totally $L\times H$ heads) independently predict their alignments, some outlier\footnote{Outlier heads mean that most of the heads are aligned in the front position, while some individual heads are aligned to the farther position, which requires the model to wait until the farthest aligned word is received, causing unnecessary latency.} heads will cause unnecessary latency \cite{Ma2019a,dualpath}.

Therefore, to better adapt to multi-head attention and capture alignments, for each decoder layer, $H$ heads in GMA jointly predict the aligned source position and share it among $H$ heads, while the predicted alignments in each decoder layer still remain independent. Since the output position (determined by predicted alignments) in each layer may be different, the model starts translating after reaching the furthest one. We will compare the performance of different sharing settings in Sec.\ref{sec:share}.

\section{Related Work}

A reasonable policy is the key to the SiMT performance. Early policies used segmented translation \cite{bangalore-etal-2012-real,Cho2016,siahbani-etal-2018-simultaneous}. \citet {gu-etal-2017-learning} used reinforcement learning to train an agent to decide read/write. \citet {Alinejad2019} added a predict operation to the agent based on \citet {gu-etal-2017-learning}.

Recent SiMT policies fall into fixed and adaptive. For fixed policy, \citet {dalvi-etal-2018-incremental} proposed STATIC-RW, which alternately read and write $RW$ words after reading $S$ words. \citet {ma-etal-2019-stacl} proposed a wait-k policy, which translates after lagging $k$ source words. \citet{multipath} enhanced wait-k policy by sampling different $k$. \citet{future-guided} proposed future-guide training for wait-k policy. \citet{han-etal-2020-end} applied meta-learning in wait-k. \citet{zhang-feng-2021-icts} proposed a char-level wait-k policy. \citet{zhang-feng-2021-universal} proposed mixture-of-experts wait-k policy.

For adaptive policy, \citet{Zheng2019b} trained an agent with golden READ/WRITE actions generated by rules. \citet {Zheng2019a} added a ``delay'' token to read source words. \citet {Arivazhagan2019} proposed MILk, using a Bernoulli variable to determine READ/WRITE. \citet {Ma2019a} proposed MMA to implement MILK on Transformer. \citet{bahar-etal-2020-start} and \citet{wilken-etal-2020-neural} used the external ground-truth alignments to train the policy. \citet{zhang-etal-2020-learning-adaptive} proposed an adaptive segmentation policy MU for SiMT. \citet{liu-etal-2021-cross} proposed cross-attention augmented transducer for SiMT. \citet{alinejad-etal-2021-translation} introduced an full-sentence model to generate a ground-truth action sequence and accordingly train a SiMT policy. \citet{miao-etal-2021-generative} proposed a generative SiMT policy.

Previous methods often neglected to jointly model alignments with translation, and meanwhile introduce additional loss functions to control the latency. However, GMA jointly learns alignment and translation, and thereby controls the latency through a simple Gaussian prior probability.

\section{Experiments}

\subsection{Datasets}

We evaluate GMA on the following public datasets.

\textbf{IWSLT15\footnote{\url{nlp.stanford.edu/projects/nmt/}} English $\!\rightarrow \!$ Vietnamese (En$\rightarrow$Vi)} (133K pairs) \cite{iwslt2015} We use TED tst2012 as validation set (1553 pairs) and TED tst2013 as test set (1268 pairs). Following the previous setting \cite{LinearTime,Ma2019a}, we replace tokens that the frequency less than 5 by $\left \langle unk \right \rangle$, and the vocabulary sizes are 17K and 7.7K for English and Vietnamese respectively.

\textbf{WMT15\footnote{\url{www.statmt.org/wmt15/translation-task}} German $\!\rightarrow\! $ English (De$\rightarrow$En)} (4.5M pairs) Following \citet{ma-etal-2019-stacl}, \citet{Arivazhagan2019} and \citet{Ma2019a}, we use newstest2013 as validation set (3000 pairs) and newstest2015 as test set (2169 pairs). BPE \cite{sennrich-etal-2016-neural} was applied with 32K merge operations and the vocabulary is shared across languages.

\subsection{Systems Setting}
We conduct experiments on the following systems.

\begin{figure*}[t]
\centering
\subfigure[En$\rightarrow$Vi, Transformer-Small]{
\includegraphics[width=1.9in]{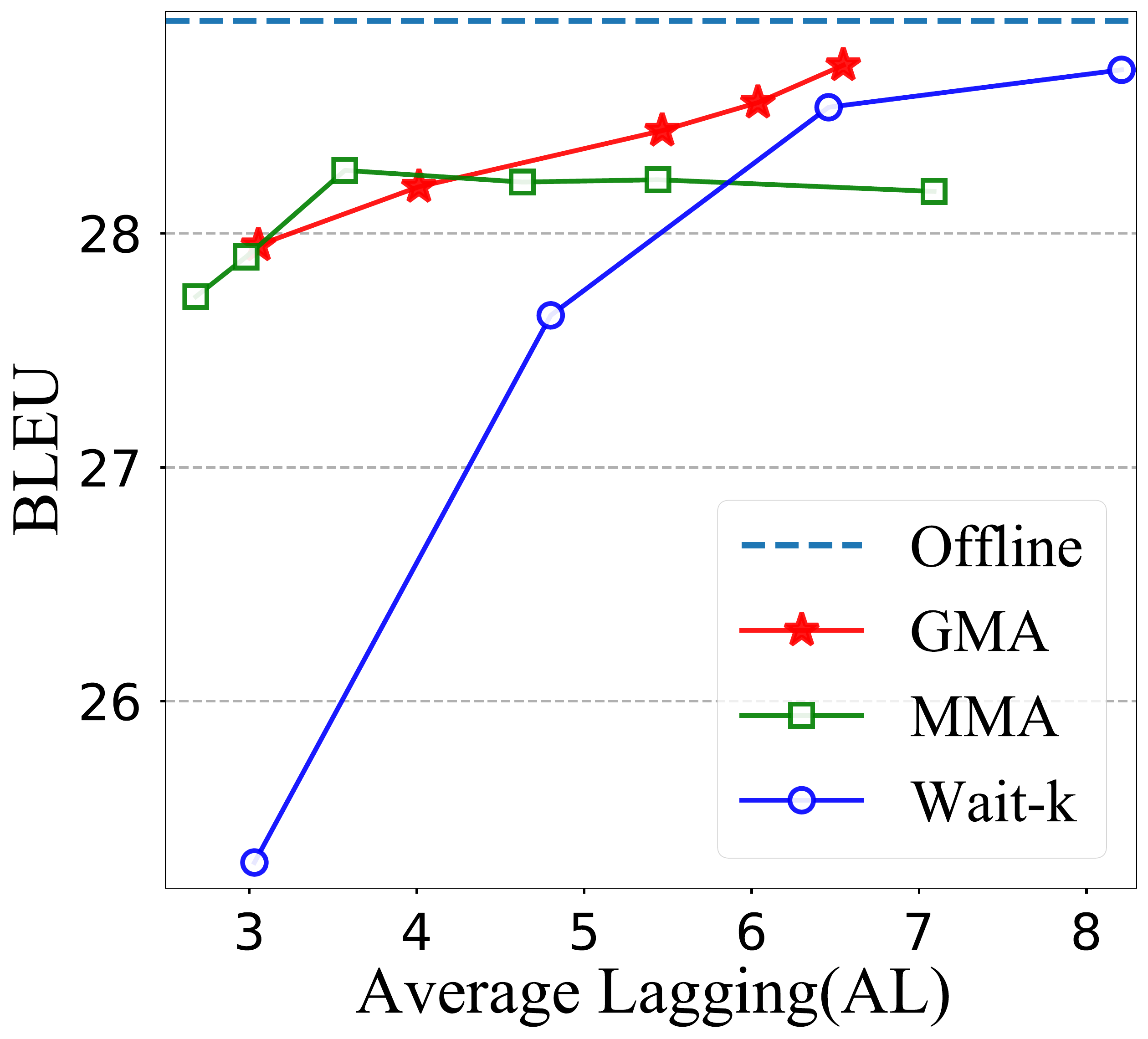}
}
\subfigure[De$\rightarrow$En, Transformer-Base]{
\includegraphics[width=1.9in]{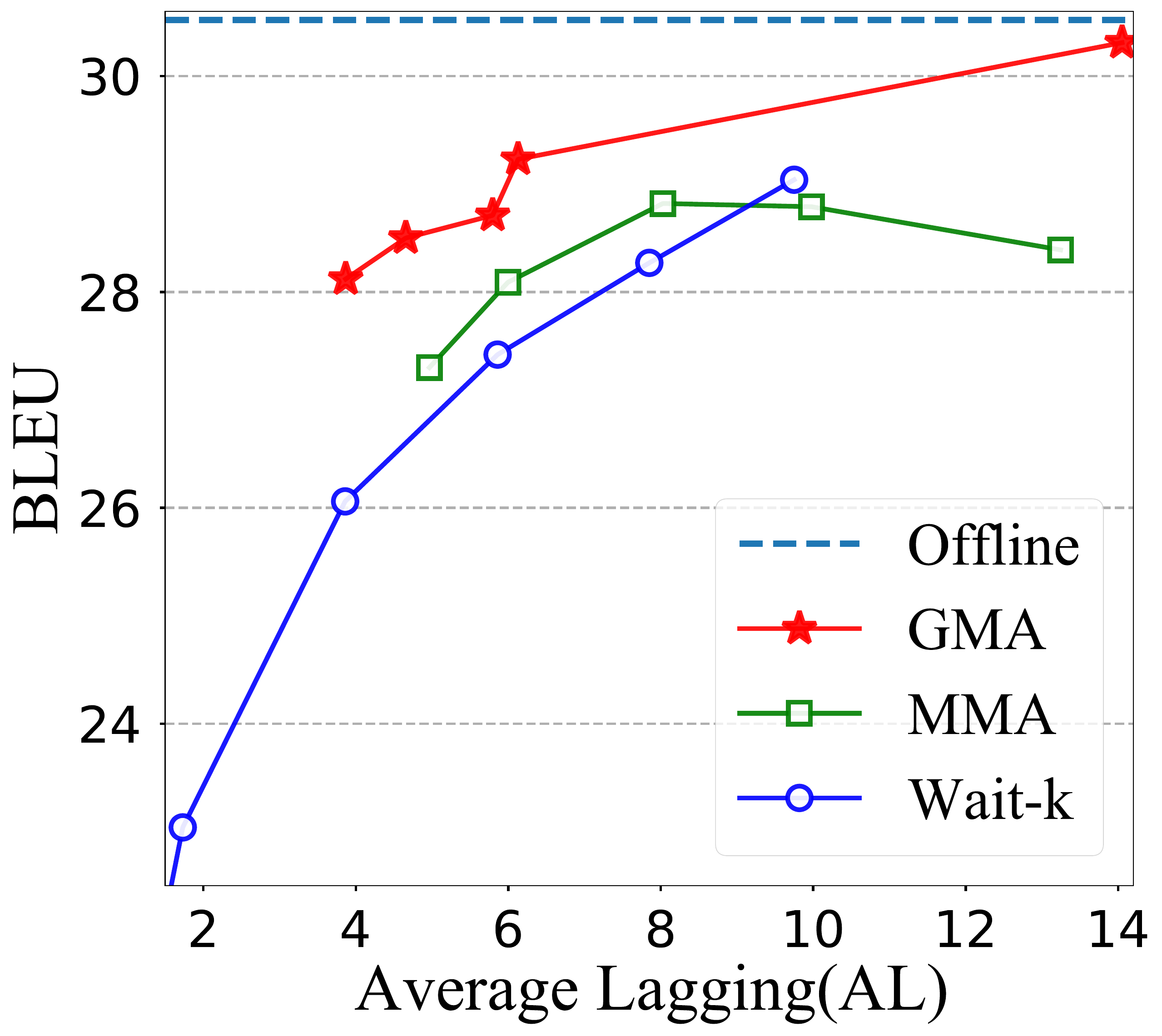}
}
\subfigure[De$\rightarrow$En, Transformer-Big]{
\includegraphics[width=1.9in]{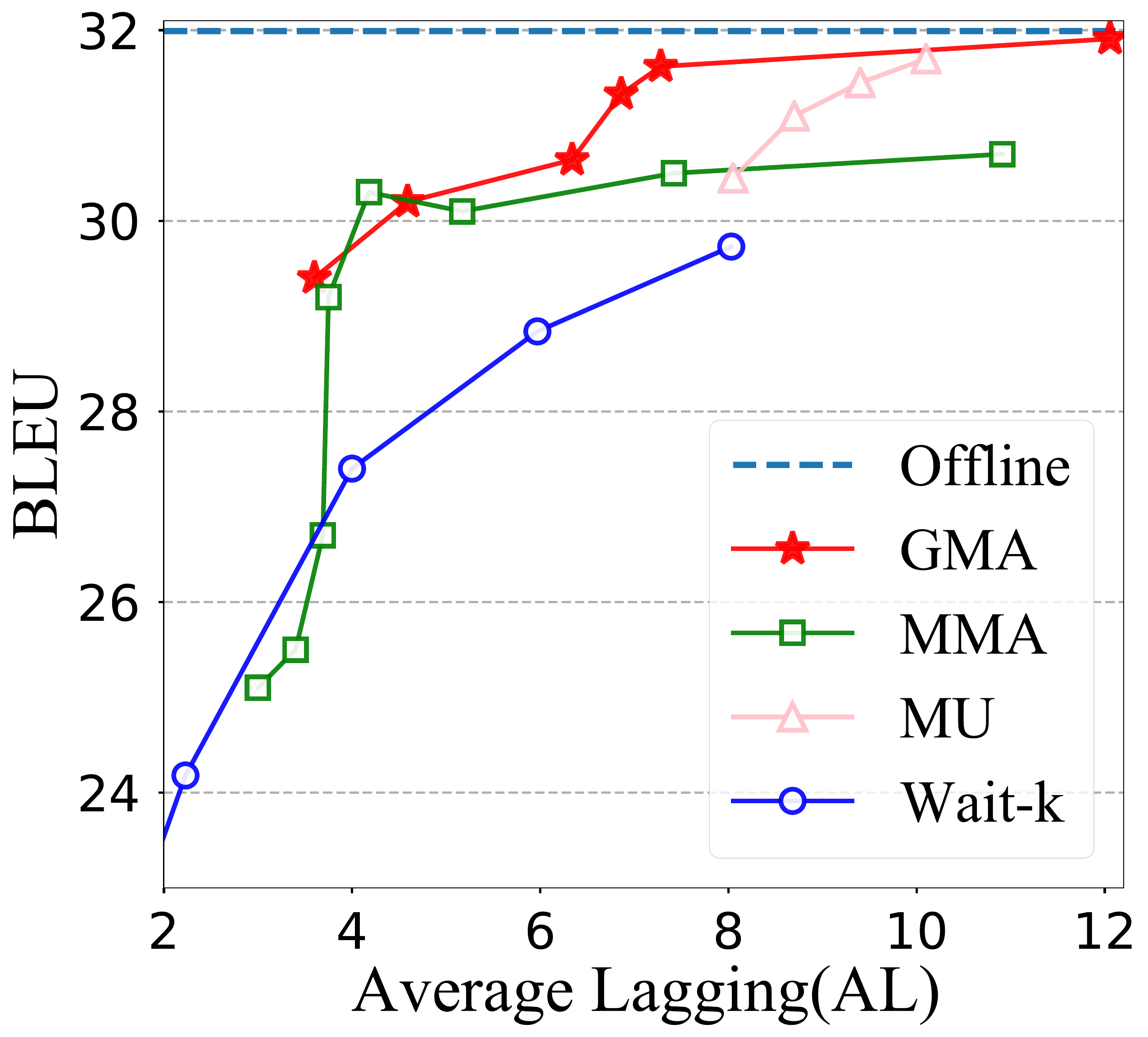}
}

\caption{Translation quality (BLEU) against latency (AL) on the En$\rightarrow$Vi(Small), De$\rightarrow$En(Base) and De$\rightarrow$En(Big), showing the results of GMA, Wait-k, MU, the SOTA adaptive policy MMA and offline model.}
\label{main}
\end{figure*}

{\bf Offline}  Conventional Transformer \cite{NIPS2017_7181} model for full-sentence translation.

{\bf Wait-k} Wait-k policy proposed by \citet{ma-etal-2019-stacl}, the most widely used fixed policy with strong performance and simple structure, which first waits for $k$ source words and then translates a target word and waits for a source word alternately.

{\bf {MU}} A segmentation policy proposed by \citet{zhang-etal-2020-learning-adaptive}, which classify whether the current inputs is a complete meaning unit (MU), and then fed MU into the full-sentence MT model for translation until generating $\left \langle \mathrm{EOS} \right \rangle$. We compare our method with MU on De$\rightarrow$En(Big) since they report their results on De$\rightarrow$En with Transformer-Big.

{\bf {MMA}\footnote{\url{github.com/pytorch/fairseq/tree/master/examples/simultaneous_translation}}} Monotonic multi-head attention (MMA) proposed by \cite{Ma2019a}, the state-of-the-art adaptive policy. At each step, MMA predicts a Bernoulli variable to decide whether to start translating or wait for the next source token.

{\bf GMA} Proposed method in Sec.\ref{sec:method}.

The implementations of all systems are adapted from Fairseq Library \cite{ott-etal-2019-fairseq} based on Transformer \cite{NIPS2017_7181}. The setting is the same as \citet {Ma2019a}. For En$\rightarrow$Vi, we apply Transformer-small (6 layers, 4 heads per layer). For De$\rightarrow$En, we apply Transformer-Base (6 layers, 8 heads per layer) and Transformer-Big (6 layers, 16 heads per layer). We evaluate these systems with BLEU \cite{papineni-etal-2002-bleu} for translation quality and Average Lagging (AL) \cite{ma-etal-2019-stacl} for latency. Average lagging is currently the most widely used latency metric, which evaluates the number of words lagging behind the ideal policy. Given $g\left ( i \right )$, AL is calculated as:
\begin{align}
    \mathrm{AL}&=\frac{1}{\tau }\sum_{i=1}^{\tau}g\left ( i \right )-\frac{i-1}{\left | \mathbf{y} \right |/\left | \mathbf{x} \right |}\\
\mathrm{where}\;\;\;  \tau &=\underset{i}{\mathrm{argmax}}\left ( g\left ( i \right )= \left | \mathbf{x} \right |\right )
\end{align}
where $\left | \mathbf{x} \right |$ and $\left | \mathbf{y} \right |$ are the length of the source sentence and target sentence respectively. 

\subsection{Main Results}

We compared GMA with the Wait-k, MU and MMA, the current best representative of fixed policy, segmentation policy and adaptive policy respectively, and plot latency-quality curves in Figure \ref{main}, where GMA curve is drawn with various $\delta$ (in Eq.(\ref{eq9})),  Wait-k curve is drawn with various lagging numbers $k$, MU curve is drawn with various classification thresholds of the meaning unit, MMA curve is drawn with various latency loss weights $\lambda$.

Compared with Wait-k, GMA has a significant improvement, since Wait-k ignores the alignments and thus the target word may be forced to be translated before receiving its aligned source word, which seriously affects the translation quality. Compared with MMA and MU, our method achieves better performance under most latency levels. Since MU first segments the source sentence based on the meaning unit, and then translates each segment with the full-sentence MT model, MU performed particularly well under high latency, but meanwhile it is difficult to extend to lower latency. 

Compared with the SOTA adaptive policy MMA, GMA has stable performance and simpler training method. MMA introduces two additional loss functions to control the latency, and meanwhile applies the expectation training to train Bernoulli variables \cite{Ma2019a}. 
GMA successfully balances the translation quality and latency without any additional loss function. Owing to the proposed Gaussian prior probability centered on predicted alignments, the source words far away from the aligned position get less Gaussian prior, so that the model is forced to move the aligned position close to the most informative source word, thereby capturing the alignments and controlling the latency. With GMA, the translation quality and latency are integrated into a unified manner and jointly optimized without any additional loss function.

\section{Analysis}

We conduct extensive analyses to study the specific improvements of GMA. Unless otherwise specified, all the results are reported on De$\rightarrow$En(Base).

\subsection{Ablation Study}
\label{sec:ab}

\begin{table}[t]
\centering
\begin{tabular}{L{4cm}|cc} \hlinew{0.7pt}
\textbf{Variants}     & \textbf{AL}       & \textbf{BLEU}     \\\hline
\multicolumn{3}{c}{\textit{\textbf{Aligned Source Position}}} \\\hline
Incremental step      & \textbf{4.66}     & \textbf{28.50}    \\ \hdashline
Absolute position     & 7.33              & 25.61            \\\hline
\multicolumn{3}{c}{\textit{\textbf{Prior Probability}}}       \\\hline
Gaussian ( $\sigma_{i}=p_{i}/2$ )             & \textbf{4.66}     & \textbf{28.50}    \\ \hdashline
$\;\;\;\;\;\;-\;\sigma_{i}=p_{i}/1$     &   6.87                &     28.96              \\
$\;\;\;\;\;\;-\;\sigma_{i}=p_{i}/3$     &   4.55                &     27.61              \\
$\;\;\;\;\;\;-$ Predicted $\sigma_{i}$     &  5.34                 &       27.12            \\
Laplace               & 8.14              & 29.19             \\
Linear                & $\!\!\!$12.83             & 27.86             \\
None                  & 1.48              & 20.84             \\\hlinew{0.7pt}
\end{tabular}
\caption{Ablation studies on modeling aligned position and the proposed prior probability, with $\delta\!=\!1$.}
\label{ablation}
\end{table}

We conduct sufficient ablation studies on the method of modeling aligned position and the proposed prior probability in Table \ref{ablation}. For modeling aligned source position, predicting incremental step performs better. Since the complete length of the streaming inputs in SiMT is unknown, predicting absolute position easily exceeds the source length, resulting in unnecessary latency. In practice, the value range of incremental step is more regular than absolute position and thus easier to learn.

Among the prior probabilities of different distributions, the Gaussian distribution performs best. The Laplace distribution is more fat-tailed than the Gaussian distribution (i.e., more prior probability on the position that far away from the aligned position), resulting in learning a later aligned position and higher latency. The linear distribution performs worse since its attenuation with distance is smoother. When removing the prior probability, since the parameters to predict alignments do not get the back-propagation gradient, the model cannot learn the alignments at all, resulting in very low latency and poor translation quality. Focusing on the best performing Gaussian prior probability, when $\sigma_{i}=p_{i}/1$ (the attenuation degree is small), the predicted aligned position is much later and the latency is higher. when $\sigma_{i}=p_{i}/3$, the translation quality declines since the prior   of some source words far away from the aligned position is too small. When the $\sigma_{i}$ is predicted, some small predicted $\sigma_{i}$ will make the prior probability of the distant source words almost 0, resulting in poor translation quality. In comparison, the Gaussian prior ($\sigma_{i}=p_{i}/2$) we proposed not only guarantees a certain attenuation degree, but also assigns the furthest source word some prior probability.

\subsection{Effect of Sharing Alignments}
\label{sec:share}
\begin{table}[t]
\centering
\begin{tabular}{L{2.8cm}|C{1.55cm}|C{0.4cm}C{1cm}} \hlinew{0.7pt}
                   & \begin{tabular}[c]{@{}c@{}}$\!\!$\textbf{\#}\textbf{Predicted} \\ $\!\!$\textbf{Alignments}\end{tabular} & \textbf{AL}$\!\!\!\!$   & \textbf{BLEU}  \\ \hline
$\!\!$All independent$\!\!$    & $6 \times 8=48\!\!\!$                                                                 & 7.85 & 29.18 \\
$\!\!$Share among heads$\!\!$  & $6 \times 1=6$                                                                     & 4.66 & 28.50 \\
$\!\!$Share among layers$\!\!$ & $1 \times 8=8$                                                                 & 4.46 & 27.82 \\
$\!\!$Share all          & $1 \times 1=1$                                                                    & 3.07 & 27.26 \\\hlinew{0.7pt}
\end{tabular}
\caption{The performance of different alignments sharing settings with $\delta\!=\!1$.}
\label{share}
\end{table}

When integrated into multi-head attention, to reduce the overall latency of the model, GMA shares the predicted alignments among $H$ heads in each layer. Table \ref{share} reports the performance of sharing alignments in different parts (all independent, among heads, among layers or among all).

`All independent' achieves the best translation quality and also brings a higher latency, as the overall latency of the model is determined by the farthest one among all predicted positions. Besides, `Share all' gets the lowest latency but loses translation quality. Comparing `Share among heads' and `Share among layers', sharing among heads performs better, which is in line with the previous conclusion that there are obvious differences between the alignments in each decoder layer \cite{garg-etal-2019-jointly,voita-etal-2019-analyzing,li-etal-2019-word}.

\subsection{Statistics of Incremental Steps}

\begin{figure}[t]
\centering
\includegraphics[width=3in]{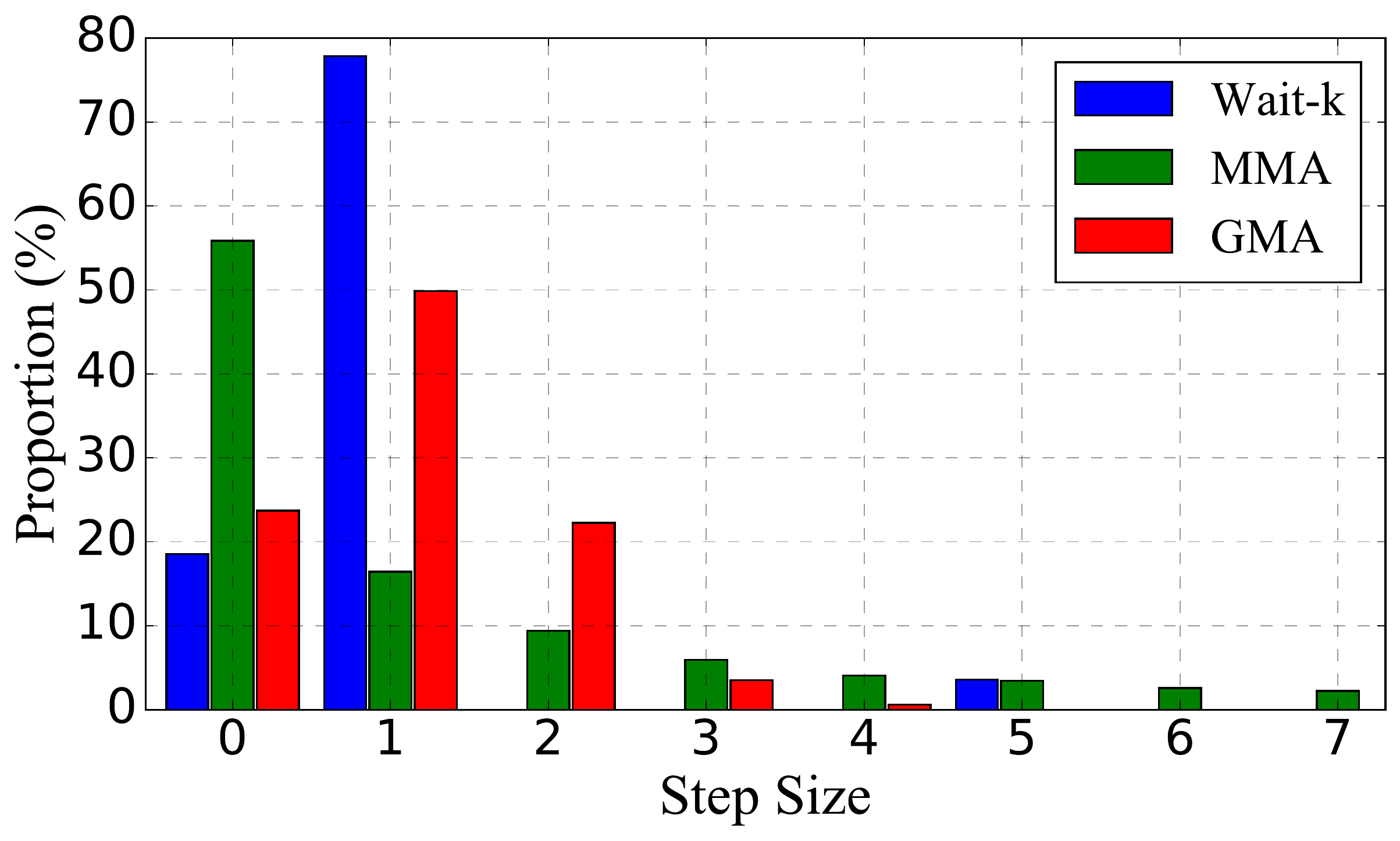}
\caption{The distribution of the waiting source number between two adjacent outputs, on Wait-k ($k\!=\!5$), MMA ($\lambda\!=\!0.4$) and GMA ($\delta\!=\!1$) with similar latency.}
\label{step}
\end{figure}

Unlike MMA predicting the READ/WRITE action, GMA directly predicts the incremental step between adjacent target outputs. To analyze the advantages of modeling the incremental step, we show the distribution of step size (i.e., the number of waiting source words between two adjacent outputs) in Figure \ref{step}, where we select the SiMT models with the similar latency (AL $\approx $ 4.5).


For Wait-k, the step size is blunt and there are only three cases, which occur before the start of translation (step size$=\!k$: first lag $k$ words), during translation (step size$=\!1$: wait and output one word alternately.), and after the end of the source inputs (step size$=\!0$: output the translation at one time). The proposed GMA and MMA have obvious differences in the step size distribution. The step size distribution of GMA is more even, only distributed between 0$\sim $4, which shows that each source segment is shorter and thus the translating process is more streaming. The step size of MMA is more widely distributed, most of which are 0 (consecutively output target words), and there is also a large proportion of step size greater than 5, which shows that MMA tends to consecutively wait for more source words and then output more target words, resulting in longer source segments. Therefore, although GMA and MMA have similar latency (AL), their performance in real applications is different, where the translation process of GMA is more streaming, while MMA is more segmented. 

\begin{figure}[t]
\centering
\includegraphics[width=2.7in]{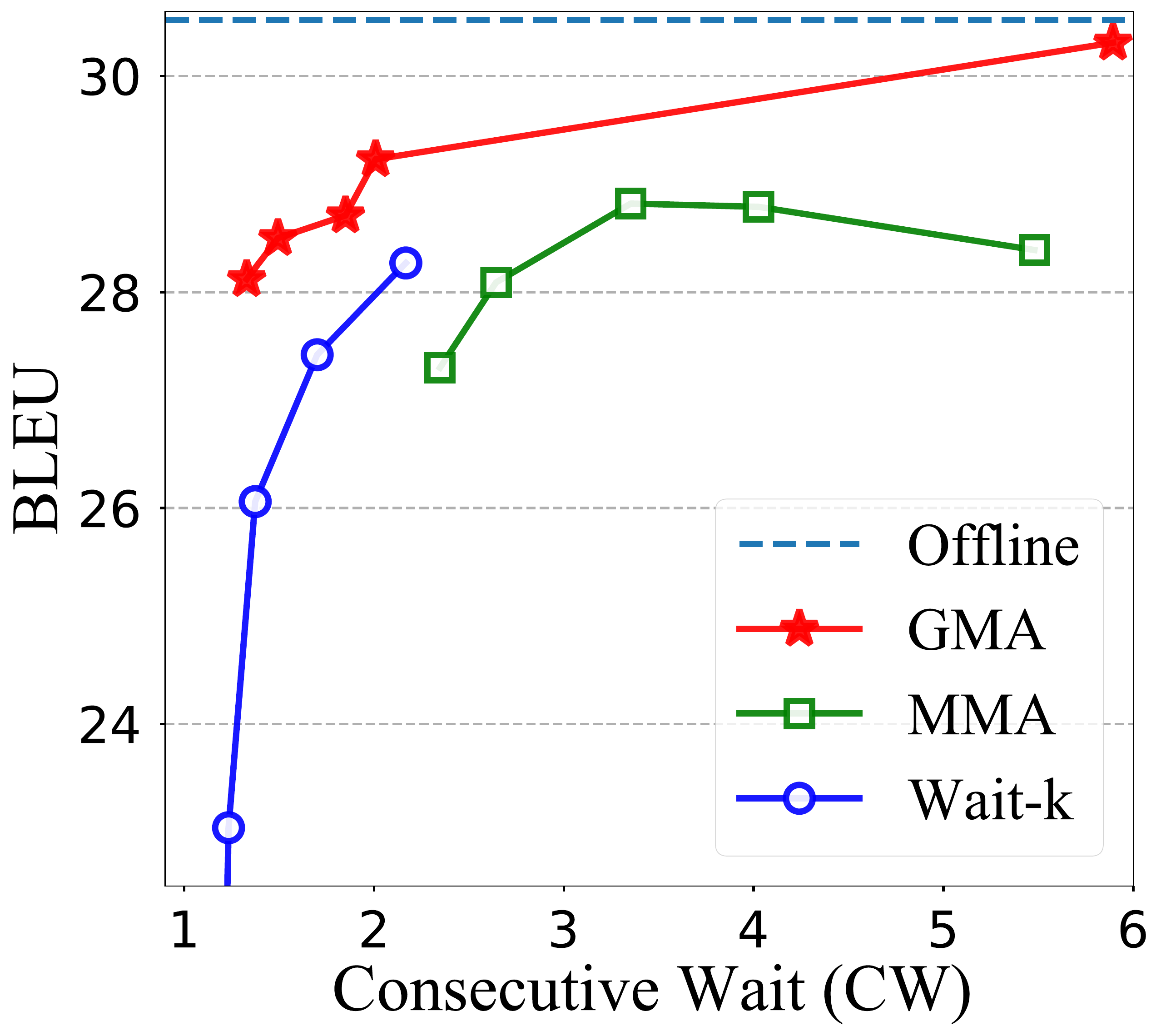}
\caption{Translation quality against latency (CW), where CW reflects the streaming degree of SiMT.}
\label{cw}
\end{figure}

Furthermore, to more accurately evaluate the streaming degree of the translation process, we apply Consecutive Wait (CW) \cite{gu-etal-2017-learning} as the latency metric to evaluate the systems. Consecutive wait evaluates the number of source words waited between two target words, which reflects the streaming degree of SiMT. Given $g\left ( i \right )$, CW is calculated as:
\begin{gather}
    \mathrm{CW}=\frac{\sum_{i=1}^{\left | \mathbf{y} \right |} (g(i)-g(i-1))}{\sum_{i=1}^{\left | \mathbf{y} \right |}\mathbbm{1}_{g(i)-g(i-1)>0}}
\end{gather}
where $\mathbbm{1}_{g(i)-g(i-1)>0}=1$ counts the number of $g(i)-g(i-1)>0$. In other words, CW measures the average source segment length (the best case is 1 for word-by-word streaming translation and the worst case is $\left | \mathbf{x} \right |$ for full-sentence MT), where the smaller the CW, the shorter the average source segment, and the translation is more streaming.

As shown in Figure \ref{cw}, GMA gets much smaller CW scores than MMA, where the average source segment length is about 2 words, which shows that GMA achieves more streaming translation than the previous adaptive policy.

\begin{figure}[t]
\centering
\includegraphics[width=3in]{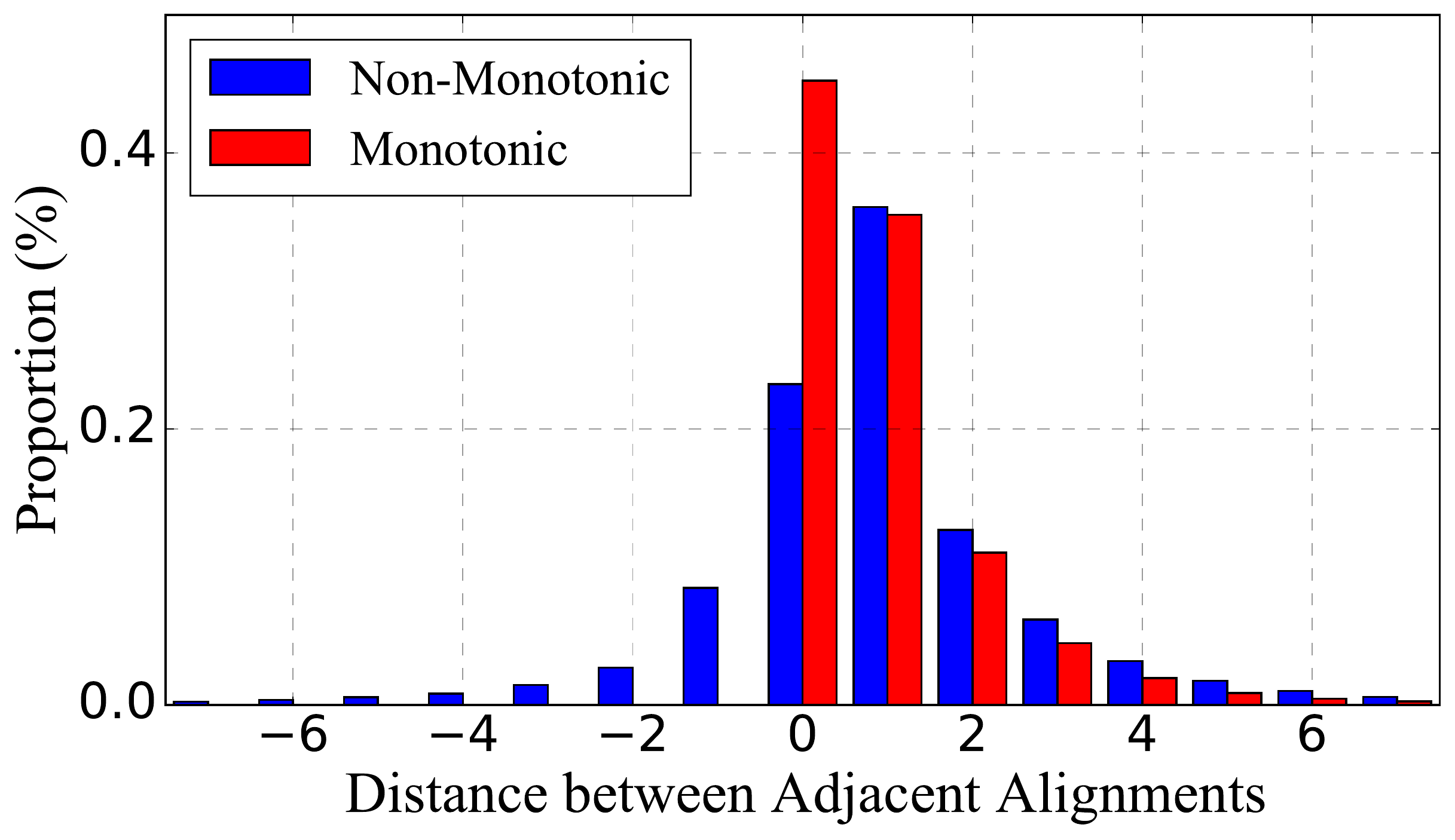}
\caption{Distribution of distances between adjacent alignments (ground-truth) under monotonic and non-monotonic alignment setting.}
\label{align_step}
\end{figure}

Additionally, although the alignments between the two languages is not necessarily monotonic, we require the predicted incremental step $\Delta p_{i} \geq 0$ to guarantee the model performs READ/WRITE monotonically. This is due to two considerations. First, we don't want to waste any useful source content, i.e., to avoid the model moving $ p_{i}$ to the previous position and thereby ignoring some received source information caused by $\Delta p_{i} < 0$. Second, we argue that monotonic alignments are more friendly to SiMT learning. We use \texttt{fast-align}\footnote{\url{https://github.com/clab/fast_align}} \cite{dyer-etal-2013-simple} to generate the ground-truth aligned source position of $i^{th}$ target token, denoted as $A_{i}$, and then show the distribution of distances between adjacent alignments under monotonic and non-monotonic alignments in Figure \ref{align_step}, where `Non-Monotonic' measures $A_{i}\!-\!A_{i-1}$ and `Monotonic' measures $\mathrm{max}(A_{i}\!-\!\mathrm{max}_{j<i}A_{j},0 )$. The distance distribution between adjacent alignments under monotonic alignment is more concentrated, between 0 and 4, which is easier for the model to learn incremental step. Actually, the incremental steps predicted by GMA almost distribute between 0 and 4 (see Figure \ref{step}), which shows that GMA successfully learns the relative distance between monotonic alignments.

\subsection{Quality of Predicted Alignments}

\begin{table}[t]
\centering
\begin{tabular}{l|c|c} \hlinew{0.7pt}
\textbf{Latency} & \textbf{AER}  & \begin{tabular}[c]{@{}c@{}}\% \textbf{of ground-truth}\\ \textbf{alignments within} $g(i)$\end{tabular} \\ \hline
Low     & 0.49 & 81.00\%                                                                        \\
Mid     & 0.61 & 88.27\%                                                                        \\
High    & 0.76 & 95.58\%           \\\hlinew{0.7pt}                                                   
\end{tabular}
\caption{Alignment quality under different latency levels, where `within $g(i)$' means that starting translating after receiving the aligned source word.}
\label{align}
\end{table}

To evaluate the quality of the aligned source position $p_{i}$ predicted by GMA, we measure the alignment accuracy on the RWTH De$\rightarrow$En alignment dataset \footnote{\url{https://www-i6.informatik.rwth-aachen.de/goldAlignment/}}, whose reference alignment was manually annotated by experts \cite{liu-etal-2016-neural,zhang-feng-2021-modeling-concentrated}. As shown in Table \ref{align}, we sample one decoder layer to calculate the alignments error rate (AER) \cite{vilar-etal-2006-aer}, and meanwhile count how many ground-truth aligned source words are located before the output position $g(i)$ (i.e., translate after receiving the aligned source word).

GMA achieves good alignment accuracy, especially at low latency, since the model is required to output immediately after receiving the aligned source words ($\delta$ in Eq.(\ref{eq9}) is small). More importantly, most of the ground-truth alignments is within $g(i)$, showing that GMA guarantees that in most cases, the model starts translating a target word after receiving its aligned source words, which is beneficial to translation quality.

\begin{figure}[t]
\includegraphics[width=3.02in]{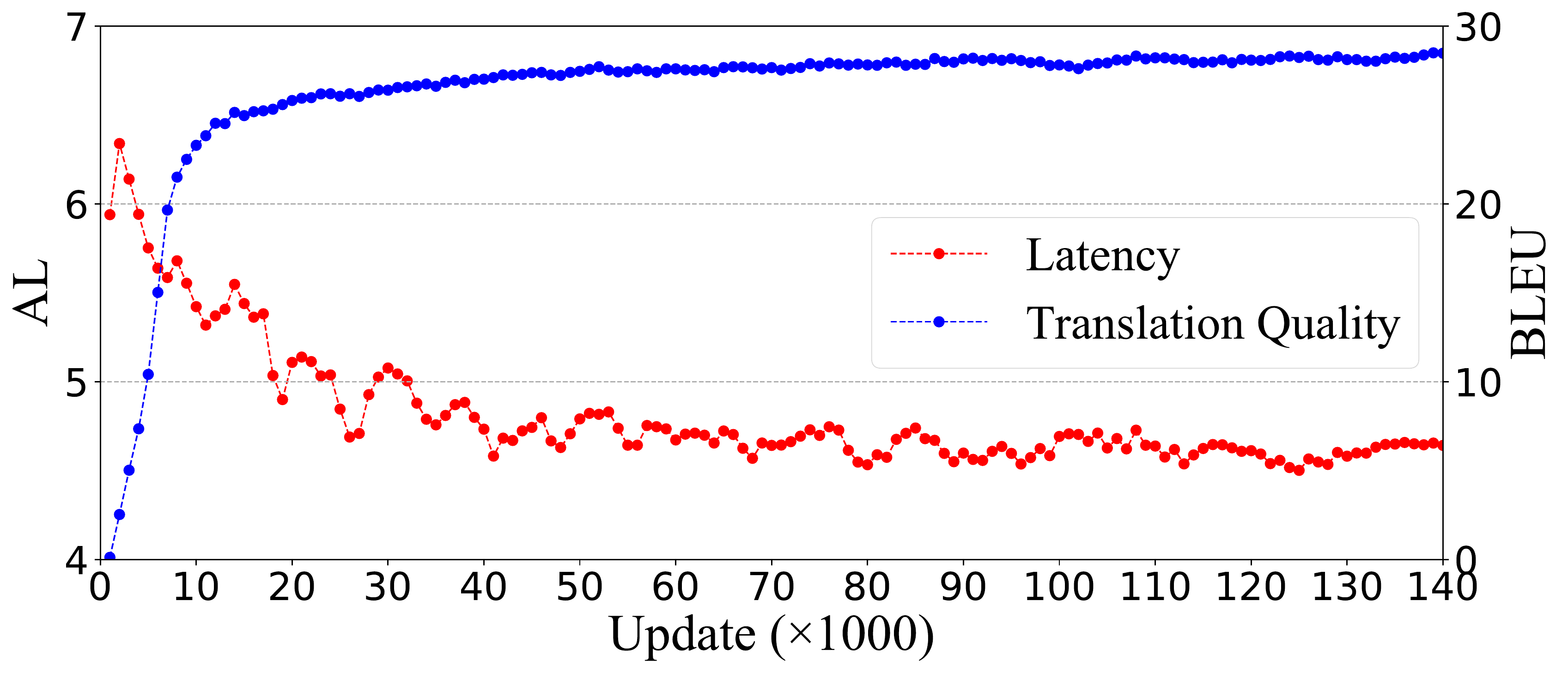}
\caption{The learning curve of translation quality (BLEU) and latency (AL) during training, with $\delta\!=\!1$.}
\label{learn}
\end{figure}

\begin{figure*}[t]
\centering
\subfigure[An easy case with more monotonic alignments]{
\includegraphics[width=6.23in]{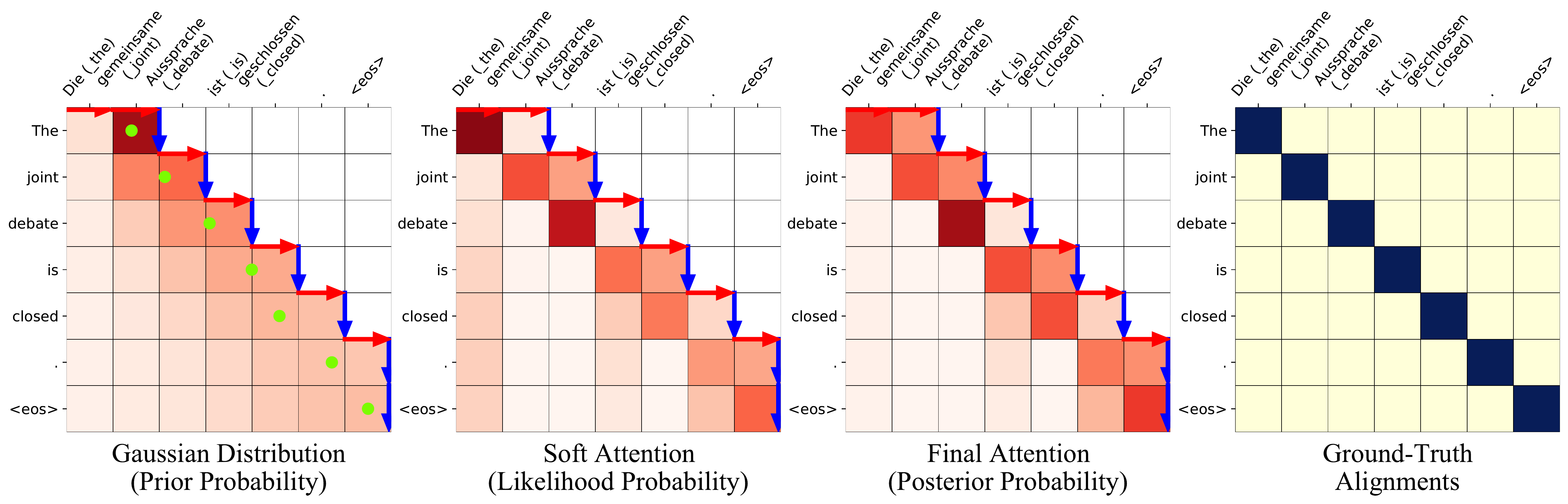}
}
\subfigure[A hard case with more complex alignments]{
\includegraphics[width=6.23in]{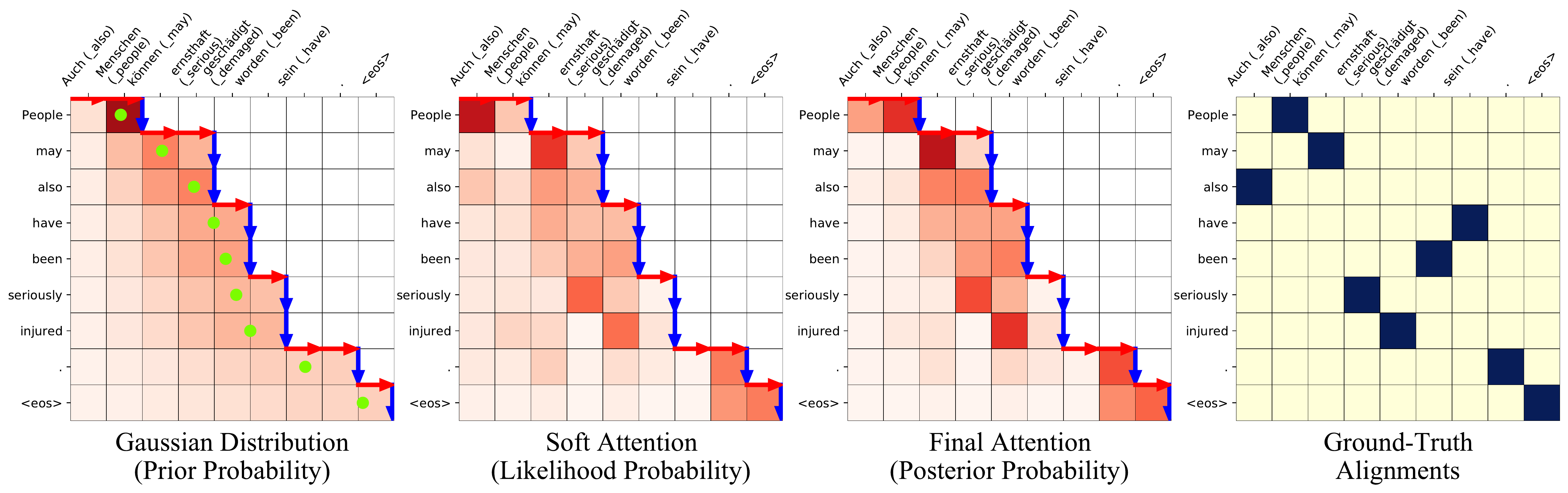}
}

\caption{Attention visualization of GMA on De$\rightarrow$En SiMT with $\delta\!=\!1$. The shade of the color indicates the attention weight. `\textcolor{green}{$\bullet $}': the predicted aligned source position $p_{i}$ (mean of Gaussian distribution), note that $p_{i}$ is a float number. `\textcolor{black}{$\blacksquare$}': the ground-truth alignments. `\textcolor{red}{$\rightarrow$}': wait for a source word,  `\textcolor{blue}{$\downarrow$}': translate a target word.}
\label{vis}
\end{figure*}

\subsection{Balancing Translation and Latency}
To study how GMA learns to balance translation quality and latency without any additional loss function during training, we draw a learning curve for translation quality and latency in Figure \ref{learn}. 

Initially, the high latency indicates that the model first moves the predicted aligned position to a further position, to learn the translation by seeing more source words. Then, as the number of training steps increases, the translation quality improves and the latency gradually decreases, which shows that for better translation, the model moves the predicted aligned source position to a more appropriate position due to the introduced Gaussian prior probability. Overall, for better translation, GMA constantly adjusts the predicted aligned source position to a suitable position and thereby controls the latency, which is completely different from the previous method of introducing the additional latency loss to constrain the latency with translation \cite{Ma2019a,miao-etal-2021-generative}.

\subsection{Characteristics of Attention in GMA}

We explore the characteristics of GMA by visualizing the attention distributions in Figure \ref{vis}. We show two cases with the alignments of different difficulty levels, where the reverse orders in alignments are considered as a major challenge for SiMT \cite{ma-etal-2019-stacl,zhang-feng-2021-universal}.

For more monotonic alignments, GMA can predict the aligned source position well and output the target word after receiving the aligned word. Meanwhile, due to the characteristics of Gaussian distribution, GMA can also avoid focusing too much on source words in the front position, and strengthen the attention on the newly received source words, which is proved to be beneficial to SiMT performance \cite{elbayad-etal-2020-online,laf}. For more complex alignments, the aligned position predicted by GMA is close to the ground-truth alignments, so that GMA starts translating after receiving most of aligned words. Besides, GMA learns some implicit prediction ability, e.g., before receiving ``\textit{worden sein}'', GMA generates the correct translation ``\textit{have been}'' based on the context. We consider this is because the predicted alignments during training are monotonic due to the incremental step, where modeling monotonic alignments forces the model to learn the correct translation from the incomplete source and previous outputs \cite{ma-etal-2019-stacl}.

\section{Conclusion}
In this paper, we propose the Gaussian multi-head attention to develop a SiMT policy which starts translating a target word after receiving its aligned source word. 
Experiments and analyses show that our method achieves promising results on performance, alignments quality and streaming degree.

\section*{Acknowledgements}
We thank all the anonymous reviewers for their insightful and valuable comments.  This work was supported by National Key R\&D Program of China (NO. 2017YFE0192900).

\bibliography{anthology,custom}

\begin{thebibliography}{40}
\expandafter\ifx\csname natexlab\endcsname\relax\def\natexlab#1{#1}\fi

\bibitem[{Alinejad et~al.(2021)Alinejad, Shavarani, and
  Sarkar}]{alinejad-etal-2021-translation}
Ashkan Alinejad, Hassan~S. Shavarani, and Anoop Sarkar. 2021.
\newblock \href {https://aclanthology.org/2021.emnlp-main.130}
  {Translation-based supervision for policy generation in simultaneous neural
  machine translation}.
\newblock In \emph{Proceedings of the 2021 Conference on Empirical Methods in
  Natural Language Processing}, pages 1734--1744, Online and Punta Cana,
  Dominican Republic. Association for Computational Linguistics.

\bibitem[{Alinejad et~al.(2018)Alinejad, Siahbani, and Sarkar}]{Alinejad2019}
Ashkan Alinejad, Maryam Siahbani, and Anoop Sarkar. 2018.
\newblock \href {https://doi.org/10.18653/v1/D18-1337} {Prediction improves
  simultaneous neural machine translation}.
\newblock In \emph{Proceedings of the 2018 Conference on Empirical Methods in
  Natural Language Processing}, pages 3022--3027, Brussels, Belgium.
  Association for Computational Linguistics.

\bibitem[{Arivazhagan et~al.(2019)Arivazhagan, Cherry, Macherey, Chiu, Yavuz,
  Pang, Li, and Raffel}]{Arivazhagan2019}
Naveen Arivazhagan, Colin Cherry, Wolfgang Macherey, Chung-Cheng Chiu, Semih
  Yavuz, Ruoming Pang, Wei Li, and Colin Raffel. 2019.
\newblock \href {https://doi.org/10.18653/v1/P19-1126} {Monotonic infinite
  lookback attention for simultaneous machine translation}.
\newblock In \emph{Proceedings of the 57th Annual Meeting of the Association
  for Computational Linguistics}, pages 1313--1323, Florence, Italy.
  Association for Computational Linguistics.

\bibitem[{Arthur et~al.(2021)Arthur, Cohn, and
  Haffari}]{arthur-etal-2021-learning}
Philip Arthur, Trevor Cohn, and Gholamreza Haffari. 2021.
\newblock \href {https://aclanthology.org/2021.eacl-main.233} {Learning coupled
  policies for simultaneous machine translation using imitation learning}.
\newblock In \emph{Proceedings of the 16th Conference of the European Chapter
  of the Association for Computational Linguistics: Main Volume}, pages
  2709--2719, Online. Association for Computational Linguistics.

\bibitem[{Bahar et~al.(2020)Bahar, Wilken, Alkhouli, Guta, Golik, Matusov, and
  Herold}]{bahar-etal-2020-start}
Parnia Bahar, Patrick Wilken, Tamer Alkhouli, Andreas Guta, Pavel Golik, Evgeny
  Matusov, and Christian Herold. 2020.
\newblock \href {https://doi.org/10.18653/v1/2020.iwslt-1.3} {Start-before-end
  and end-to-end: Neural speech translation by {A}pp{T}ek and {RWTH} {A}achen
  {U}niversity}.
\newblock In \emph{Proceedings of the 17th International Conference on Spoken
  Language Translation}, pages 44--54, Online. Association for Computational
  Linguistics.

\bibitem[{Bangalore et~al.(2012)Bangalore, Rangarajan~Sridhar, Kolan, Golipour,
  and Jimenez}]{bangalore-etal-2012-real}
Srinivas Bangalore, Vivek~Kumar Rangarajan~Sridhar, Prakash Kolan, Ladan
  Golipour, and Aura Jimenez. 2012.
\newblock \href {https://www.aclweb.org/anthology/N12-1048} {Real-time
  incremental speech-to-speech translation of dialogs}.
\newblock In \emph{Proceedings of the 2012 Conference of the North {A}merican
  Chapter of the Association for Computational Linguistics: Human Language
  Technologies}, pages 437--445, Montr{\'e}al, Canada. Association for
  Computational Linguistics.

\bibitem[{Cettolo et~al.(2015)Cettolo, Jan, Sebastian, Bentivogli, Cattoni, and
  Federico}]{iwslt2015}
Mauro Cettolo, Niehues Jan, St{\"u}ker Sebastian, Luisa Bentivogli, R.~Cattoni,
  and Marcello Federico. 2015.
\newblock The iwslt 2015 evaluation campaign.

\bibitem[{Cho and Esipova(2016)}]{Cho2016}
Kyunghyun Cho and Masha Esipova. 2016.
\newblock \href {http://arxiv.org/abs/1606.02012} {{Can neural machine
  translation do simultaneous translation?}}

\bibitem[{Dalvi et~al.(2018)Dalvi, Durrani, Sajjad, and
  Vogel}]{dalvi-etal-2018-incremental}
Fahim Dalvi, Nadir Durrani, Hassan Sajjad, and Stephan Vogel. 2018.
\newblock \href {https://doi.org/10.18653/v1/N18-2079} {Incremental decoding
  and training methods for simultaneous translation in neural machine
  translation}.
\newblock In \emph{Proceedings of the 2018 Conference of the North {A}merican
  Chapter of the Association for Computational Linguistics: Human Language
  Technologies, Volume 2 (Short Papers)}, pages 493--499, New Orleans,
  Louisiana. Association for Computational Linguistics.

\bibitem[{Dyer et~al.(2013)Dyer, Chahuneau, and Smith}]{dyer-etal-2013-simple}
Chris Dyer, Victor Chahuneau, and Noah~A. Smith. 2013.
\newblock \href {https://www.aclweb.org/anthology/N13-1073} {A simple, fast,
  and effective reparameterization of {IBM} model 2}.
\newblock In \emph{Proceedings of the 2013 Conference of the North {A}merican
  Chapter of the Association for Computational Linguistics: Human Language
  Technologies}, pages 644--648, Atlanta, Georgia. Association for
  Computational Linguistics.

\bibitem[{Elbayad et~al.(2020{\natexlab{a}})Elbayad, Besacier, and
  Verbeek}]{multipath}
Maha Elbayad, Laurent Besacier, and Jakob Verbeek. 2020{\natexlab{a}}.
\newblock \href {https://doi.org/10.21437/Interspeech.2020-1241} {{Efficient
  Wait-k Models for Simultaneous Machine Translation}}.

\bibitem[{Elbayad et~al.(2020{\natexlab{b}})Elbayad, Ustaszewski,
  Esperan{\c{c}}a-Rodier, Brunet-Manquat, Verbeek, and
  Besacier}]{elbayad-etal-2020-online}
Maha Elbayad, Michael Ustaszewski, Emmanuelle Esperan{\c{c}}a-Rodier, Francis
  Brunet-Manquat, Jakob Verbeek, and Laurent Besacier. 2020{\natexlab{b}}.
\newblock \href {https://doi.org/10.18653/v1/2020.coling-main.443} {Online
  versus offline {NMT} quality: An in-depth analysis on {E}nglish-{G}erman and
  {G}erman-{E}nglish}.
\newblock In \emph{Proceedings of the 28th International Conference on
  Computational Linguistics}, pages 5047--5058, Barcelona, Spain (Online).
  International Committee on Computational Linguistics.

\bibitem[{Garg et~al.(2019)Garg, Peitz, Nallasamy, and
  Paulik}]{garg-etal-2019-jointly}
Sarthak Garg, Stephan Peitz, Udhyakumar Nallasamy, and Matthias Paulik. 2019.
\newblock \href {https://doi.org/10.18653/v1/D19-1453} {Jointly learning to
  align and translate with transformer models}.
\newblock In \emph{Proceedings of the 2019 Conference on Empirical Methods in
  Natural Language Processing and the 9th International Joint Conference on
  Natural Language Processing (EMNLP-IJCNLP)}, pages 4453--4462, Hong Kong,
  China. Association for Computational Linguistics.

\bibitem[{Gu et~al.(2017)Gu, Neubig, Cho, and Li}]{gu-etal-2017-learning}
Jiatao Gu, Graham Neubig, Kyunghyun Cho, and Victor~O.K. Li. 2017.
\newblock \href {https://www.aclweb.org/anthology/E17-1099} {Learning to
  translate in real-time with neural machine translation}.
\newblock In \emph{Proceedings of the 15th Conference of the {E}uropean Chapter
  of the Association for Computational Linguistics: Volume 1, Long Papers},
  pages 1053--1062, Valencia, Spain. Association for Computational Linguistics.

\bibitem[{Han et~al.(2020)Han, Zaidi, Indurthi, Lakumarapu, Lee, and
  Kim}]{han-etal-2020-end}
Hou~Jeung Han, Mohd~Abbas Zaidi, Sathish~Reddy Indurthi, Nikhil~Kumar
  Lakumarapu, Beomseok Lee, and Sangha Kim. 2020.
\newblock \href {https://doi.org/10.18653/v1/2020.iwslt-1.5} {End-to-end
  simultaneous translation system for {IWSLT}2020 using modality agnostic
  meta-learning}.
\newblock In \emph{Proceedings of the 17th International Conference on Spoken
  Language Translation}, pages 62--68, Online. Association for Computational
  Linguistics.

\bibitem[{Li et~al.(2019)Li, Li, Liu, Meng, and Shi}]{li-etal-2019-word}
Xintong Li, Guanlin Li, Lemao Liu, Max Meng, and Shuming Shi. 2019.
\newblock \href {https://doi.org/10.18653/v1/P19-1124} {On the word alignment
  from neural machine translation}.
\newblock In \emph{Proceedings of the 57th Annual Meeting of the Association
  for Computational Linguistics}, pages 1293--1303, Florence, Italy.
  Association for Computational Linguistics.

\bibitem[{Liu et~al.(2021)Liu, Du, Li, Li, and Chen}]{liu-etal-2021-cross}
Dan Liu, Mengge Du, Xiaoxi Li, Ya~Li, and Enhong Chen. 2021.
\newblock \href {https://aclanthology.org/2021.emnlp-main.4} {Cross attention
  augmented transducer networks for simultaneous translation}.
\newblock In \emph{Proceedings of the 2021 Conference on Empirical Methods in
  Natural Language Processing}, pages 39--55, Online and Punta Cana, Dominican
  Republic. Association for Computational Linguistics.

\bibitem[{Liu et~al.(2016)Liu, Utiyama, Finch, and
  Sumita}]{liu-etal-2016-neural}
Lemao Liu, Masao Utiyama, Andrew Finch, and Eiichiro Sumita. 2016.
\newblock \href {https://www.aclweb.org/anthology/C16-1291} {Neural machine
  translation with supervised attention}.
\newblock In \emph{Proceedings of {COLING} 2016, the 26th International
  Conference on Computational Linguistics: Technical Papers}, pages 3093--3102,
  Osaka, Japan. The COLING 2016 Organizing Committee.

\bibitem[{Ma et~al.(2019)Ma, Huang, Xiong, Zheng, Liu, Zheng, Zhang, He, Liu,
  Li, Wu, and Wang}]{ma-etal-2019-stacl}
Mingbo Ma, Liang Huang, Hao Xiong, Renjie Zheng, Kaibo Liu, Baigong Zheng,
  Chuanqiang Zhang, Zhongjun He, Hairong Liu, Xing Li, Hua Wu, and Haifeng
  Wang. 2019.
\newblock \href {https://doi.org/10.18653/v1/P19-1289} {{STACL}: Simultaneous
  translation with implicit anticipation and controllable latency using
  prefix-to-prefix framework}.
\newblock In \emph{Proceedings of the 57th Annual Meeting of the Association
  for Computational Linguistics}, pages 3025--3036, Florence, Italy.
  Association for Computational Linguistics.

\bibitem[{Ma et~al.(2020)Ma, Pino, Cross, Puzon, and Gu}]{Ma2019a}
Xutai Ma, Juan~Miguel Pino, James Cross, Liezl Puzon, and Jiatao Gu. 2020.
\newblock \href {https://openreview.net/forum?id=Hyg96gBKPS} {Monotonic
  multihead attention}.
\newblock In \emph{International Conference on Learning Representations}.

\bibitem[{Miao et~al.(2021)Miao, Blunsom, and
  Specia}]{miao-etal-2021-generative}
Yishu Miao, Phil Blunsom, and Lucia Specia. 2021.
\newblock \href {https://aclanthology.org/2021.emnlp-main.536} {A generative
  framework for simultaneous machine translation}.
\newblock In \emph{Proceedings of the 2021 Conference on Empirical Methods in
  Natural Language Processing}, pages 6697--6706, Online and Punta Cana,
  Dominican Republic. Association for Computational Linguistics.

\bibitem[{Ott et~al.(2019)Ott, Edunov, Baevski, Fan, Gross, Ng, Grangier, and
  Auli}]{ott-etal-2019-fairseq}
Myle Ott, Sergey Edunov, Alexei Baevski, Angela Fan, Sam Gross, Nathan Ng,
  David Grangier, and Michael Auli. 2019.
\newblock \href {https://doi.org/10.18653/v1/N19-4009} {fairseq: A fast,
  extensible toolkit for sequence modeling}.
\newblock In \emph{Proceedings of the 2019 Conference of the North {A}merican
  Chapter of the Association for Computational Linguistics (Demonstrations)},
  pages 48--53, Minneapolis, Minnesota. Association for Computational
  Linguistics.

\bibitem[{Papineni et~al.(2002)Papineni, Roukos, Ward, and
  Zhu}]{papineni-etal-2002-bleu}
Kishore Papineni, Salim Roukos, Todd Ward, and Wei-Jing Zhu. 2002.
\newblock \href {https://doi.org/10.3115/1073083.1073135} {{B}leu: a method for
  automatic evaluation of machine translation}.
\newblock In \emph{Proceedings of the 40th Annual Meeting of the Association
  for Computational Linguistics}, pages 311--318, Philadelphia, Pennsylvania,
  USA. Association for Computational Linguistics.

\bibitem[{Pukelsheim(1994)}]{2sigma}
Friedrich Pukelsheim. 1994.
\newblock \href {http://www.jstor.org/stable/2684253} {The three sigma rule}.
\newblock \emph{The American Statistician}, 48(2):88--91.

\bibitem[{Raffel et~al.(2017)Raffel, Luong, Liu, Weiss, and Eck}]{LinearTime}
Colin Raffel, Minh-Thang Luong, Peter~J. Liu, Ron~J. Weiss, and Douglas Eck.
  2017.
\newblock \href {https://proceedings.mlr.press/v70/raffel17a.html} {Online and
  linear-time attention by enforcing monotonic alignments}.
\newblock In \emph{Proceedings of the 34th International Conference on Machine
  Learning}, volume~70 of \emph{Proceedings of Machine Learning Research},
  pages 2837--2846. PMLR.

\bibitem[{Sennrich et~al.(2016)Sennrich, Haddow, and
  Birch}]{sennrich-etal-2016-neural}
Rico Sennrich, Barry Haddow, and Alexandra Birch. 2016.
\newblock \href {https://doi.org/10.18653/v1/P16-1162} {Neural machine
  translation of rare words with subword units}.
\newblock In \emph{Proceedings of the 54th Annual Meeting of the Association
  for Computational Linguistics (Volume 1: Long Papers)}, pages 1715--1725,
  Berlin, Germany. Association for Computational Linguistics.

\bibitem[{Siahbani et~al.(2018)Siahbani, Shavarani, Alinejad, and
  Sarkar}]{siahbani-etal-2018-simultaneous}
Maryam Siahbani, Hassan Shavarani, Ashkan Alinejad, and Anoop Sarkar. 2018.
\newblock \href {https://www.aclweb.org/anthology/W18-1815} {Simultaneous
  translation using optimized segmentation}.
\newblock In \emph{Proceedings of the 13th Conference of the Association for
  Machine Translation in the {A}mericas (Volume 1: Research Papers)}, pages
  154--167, Boston, MA. Association for Machine Translation in the Americas.

\bibitem[{Vaswani et~al.(2017)Vaswani, Shazeer, Parmar, Uszkoreit, Jones,
  Gomez, Kaiser, and Polosukhin}]{NIPS2017_7181}
Ashish Vaswani, Noam Shazeer, Niki Parmar, Jakob Uszkoreit, Llion Jones,
  Aidan~N Gomez, \L~ukasz Kaiser, and Illia Polosukhin. 2017.
\newblock \href
  {http://papers.nips.cc/paper/7181-attention-is-all-you-need.pdf} {Attention
  is all you need}.
\newblock In I.~Guyon, U.~V. Luxburg, S.~Bengio, H.~Wallach, R.~Fergus,
  S.~Vishwanathan, and R.~Garnett, editors, \emph{Advances in Neural
  Information Processing Systems 30}, pages 5998--6008. Curran Associates, Inc.

\bibitem[{Vilar et~al.(2006)Vilar, Popovic, and Ney}]{vilar-etal-2006-aer}
David Vilar, Maja Popovic, and Hermann Ney. 2006.
\newblock \href {https://aclanthology.org/2006.iwslt-papers.7} {{AER}: do we
  need to {``}improve{''} our alignments?}
\newblock In \emph{Proceedings of the Third International Workshop on Spoken
  Language Translation: Papers}, Kyoto, Japan.

\bibitem[{Voita et~al.(2019)Voita, Talbot, Moiseev, Sennrich, and
  Titov}]{voita-etal-2019-analyzing}
Elena Voita, David Talbot, Fedor Moiseev, Rico Sennrich, and Ivan Titov. 2019.
\newblock \href {https://doi.org/10.18653/v1/P19-1580} {Analyzing multi-head
  self-attention: Specialized heads do the heavy lifting, the rest can be
  pruned}.
\newblock In \emph{Proceedings of the 57th Annual Meeting of the Association
  for Computational Linguistics}, pages 5797--5808, Florence, Italy.
  Association for Computational Linguistics.

\bibitem[{Wilken et~al.(2020)Wilken, Alkhouli, Matusov, and
  Golik}]{wilken-etal-2020-neural}
Patrick Wilken, Tamer Alkhouli, Evgeny Matusov, and Pavel Golik. 2020.
\newblock \href {https://doi.org/10.18653/v1/2020.iwslt-1.29} {Neural
  simultaneous speech translation using alignment-based chunking}.
\newblock In \emph{Proceedings of the 17th International Conference on Spoken
  Language Translation}, pages 237--246, Online. Association for Computational
  Linguistics.

\bibitem[{Zhang et~al.(2020)Zhang, Zhang, He, Wu, and
  Wang}]{zhang-etal-2020-learning-adaptive}
Ruiqing Zhang, Chuanqiang Zhang, Zhongjun He, Hua Wu, and Haifeng Wang. 2020.
\newblock \href {https://doi.org/10.18653/v1/2020.emnlp-main.178} {Learning
  adaptive segmentation policy for simultaneous translation}.
\newblock In \emph{Proceedings of the 2020 Conference on Empirical Methods in
  Natural Language Processing (EMNLP)}, pages 2280--2289, Online. Association
  for Computational Linguistics.

\bibitem[{Zhang and Feng(2021{\natexlab{a}})}]{zhang-feng-2021-icts}
Shaolei Zhang and Yang Feng. 2021{\natexlab{a}}.
\newblock \href {https://doi.org/10.18653/v1/2021.autosimtrans-1.1} {{ICT}{'}s
  system for {A}uto{S}im{T}rans 2021: Robust char-level simultaneous
  translation}.
\newblock In \emph{Proceedings of the Second Workshop on Automatic Simultaneous
  Translation}, pages 1--11, Online. Association for Computational Linguistics.

\bibitem[{Zhang and
  Feng(2021{\natexlab{b}})}]{zhang-feng-2021-modeling-concentrated}
Shaolei Zhang and Yang Feng. 2021{\natexlab{b}}.
\newblock \href {https://doi.org/10.18653/v1/2021.findings-emnlp.121} {Modeling
  concentrated cross-attention for neural machine translation with {G}aussian
  mixture model}.
\newblock In \emph{Findings of the Association for Computational Linguistics:
  EMNLP 2021}, pages 1401--1411, Punta Cana, Dominican Republic. Association
  for Computational Linguistics.

\bibitem[{Zhang and Feng(2021{\natexlab{c}})}]{zhang-feng-2021-universal}
Shaolei Zhang and Yang Feng. 2021{\natexlab{c}}.
\newblock \href {https://aclanthology.org/2021.emnlp-main.581} {Universal
  simultaneous machine translation with mixture-of-experts wait-k policy}.
\newblock In \emph{Proceedings of the 2021 Conference on Empirical Methods in
  Natural Language Processing}, pages 7306--7317, Online and Punta Cana,
  Dominican Republic. Association for Computational Linguistics.

\bibitem[{Zhang and Feng(2022{\natexlab{a}})}]{dualpath}
Shaolei Zhang and Yang Feng. 2022{\natexlab{a}}.
\newblock Modeling dual read/write paths for simultaneous machine translation.
\newblock In \emph{Proceedings of the 60th Annual Meeting of the Association
  for Computational Linguistics}, Dublin, Ireland. Association for
  Computational Linguistics.

\bibitem[{Zhang and Feng(2022{\natexlab{b}})}]{laf}
Shaolei Zhang and Yang Feng. 2022{\natexlab{b}}.
\newblock Reducing position bias in simultaneous machine translation with
  length-aware framework.
\newblock In \emph{Proceedings of the 60th Annual Meeting of the Association
  for Computational Linguistics}, Dublin, Ireland. Association for
  Computational Linguistics.

\bibitem[{Zhang et~al.(2021)Zhang, Feng, and Li}]{future-guided}
Shaolei Zhang, Yang Feng, and Liangyou Li. 2021.
\newblock \href {https://ojs.aaai.org/index.php/AAAI/article/view/17696}
  {Future-guided incremental transformer for simultaneous translation}.
\newblock \emph{Proceedings of the AAAI Conference on Artificial Intelligence},
  35(16):14428--14436.

\bibitem[{Zheng et~al.(2019{\natexlab{a}})Zheng, Zheng, Ma, and
  Huang}]{Zheng2019b}
Baigong Zheng, Renjie Zheng, Mingbo Ma, and Liang Huang. 2019{\natexlab{a}}.
\newblock \href {https://doi.org/10.18653/v1/D19-1137} {Simpler and faster
  learning of adaptive policies for simultaneous translation}.
\newblock In \emph{Proceedings of the 2019 Conference on Empirical Methods in
  Natural Language Processing and the 9th International Joint Conference on
  Natural Language Processing (EMNLP-IJCNLP)}, pages 1349--1354, Hong Kong,
  China. Association for Computational Linguistics.

\bibitem[{Zheng et~al.(2019{\natexlab{b}})Zheng, Zheng, Ma, and
  Huang}]{Zheng2019a}
Baigong Zheng, Renjie Zheng, Mingbo Ma, and Liang Huang. 2019{\natexlab{b}}.
\newblock \href {https://doi.org/10.18653/v1/P19-1582} {Simultaneous
  translation with flexible policy via restricted imitation learning}.
\newblock In \emph{Proceedings of the 57th Annual Meeting of the Association
  for Computational Linguistics}, pages 5816--5822, Florence, Italy.
  Association for Computational Linguistics.

\end{thebibliography}
\bibliographystyle{acl_natbib}

\appendix

\section{Numerical Results}
We additionally use Consecutive Wait (CW) \cite{gu-etal-2017-learning}, Average Proportion (AP) \cite{Cho2016}, Average Lagging (AL) \cite{ma-etal-2019-stacl}, and Differentiable Average Lagging (DAL) \cite{Arivazhagan2019} to evaluate the latency of GMA, and the numerical results are shown in Table \ref{envismall}, \ref{deenbase} and \ref{deenbig}.

\newpage
\begin{table}[t]
\centering
\begin{tabular}{cccccc} \hlinew{1.2pt}
\multicolumn{6}{c}{\textbf{IWSLT15   En$\rightarrow$Vi (Small)}}                                    \\ 
\textbf{$\delta$} & \textbf{CW} & \textbf{AP} & \textbf{AL} & \textbf{DAL} & \textbf{BLEU} \\ \hline
0.9            & 1.20        & 0.65        & 3.05        & 4.08         & 27.95         \\
1.0            & 1.27        & 0.68        & 4.01        & 4.77         & 28.20         \\
2.0            & 1.49        & 0.74        & 5.47        & 6.37         & 28.44         \\
2.2            & 1.60        & 0.77        & 6.04        & 6.96         & 28.56         \\
2.5            & 1.74        & 0.78        & 6.55        & 7.55         & 28.72        \\ \hlinew{1.2pt}
\end{tabular}
\caption{Numerical results of GMA on IWSLT15 En$\rightarrow$Vi (Small).}
\label{envismall}
\end{table}

\begin{table}[t]
\centering
\begin{tabular}{cccccc} \hlinew{1.2pt}
\multicolumn{6}{c}{\textbf{WMT15   De$\rightarrow$En (Base)}}                                       \\
\textbf{$\delta$} & \textbf{CW} & \textbf{AP} & \textbf{AL} & \textbf{DAL} & \textbf{BLEU} \\ \hline
0.9            & 1.33        & 0.64        & 3.87        & 4.61         & 28.12         \\
1.0            & 1.49        & 0.67        & 4.66        & 5.56         & 28.50         \\
2.0            & 1.85        & 0.72        & 5.79        & 7.75         & 28.71         \\
2.2            & 2.01        & 0.73        & 6.13        & 8.43         & 29.23         \\
2.4            & 5.89        & 0.96        & 14.05       & 25.76        & 31.31        \\ \hlinew{1.2pt}
\end{tabular}
\caption{Numerical results of GMA on WMT15 De$\rightarrow$En (Base).}
\label{deenbase}
\end{table}

\begin{table}[t]
\centering
\begin{tabular}{cccccc} \hlinew{1.2pt}
\multicolumn{6}{c}{\textbf{WMT15   De$\rightarrow$En (Big)}}                                        \\
\textbf{$\delta$} & \textbf{CW} & \textbf{AP} & \textbf{AL} & \textbf{DAL} & \textbf{BLEU} \\ \hline
1.0            & 1.54        & 0.68        & 4.60        & 5.89         & 30.20         \\
2.0            & 1.98        & 0.74        & 6.34        & 8.18         & 30.64         \\
2.2            & 2.13        & 0.75        & 6.86        & 8.91         & 31.33         \\
2.4            & 2.28        & 0.76        & 7.28        & 9.59         & 31.62         \\
2.5            & 3.10        & 0.88        & 12.06       & 20.43        & 31.91        \\ \hlinew{1.2pt}
\end{tabular}
\caption{Numerical results of GMA on WMT15 De$\rightarrow$En (Big).}
\label{deenbig}
\end{table}

\end{document}